\documentclass[runningheads]{llncs}
\usepackage{subcaption}
\usepackage{url}
\usepackage{graphicx}
\usepackage{paralist}
\usepackage{footnote}
\usepackage{enumerate}
\usepackage[skip=4pt]{caption}

\newcommand{\etal}{\emph{et al.} }
\newcommand{\ie}{\emph{i.e.}, }
\newcommand{\eg}{\emph{e.g.}, }
\newcommand{\etc}{\emph{etc.}}

\begin{document}
\title{MoViDNN: A Mobile Platform for Evaluating Video Quality Enhancement with Deep Neural Networks}
\titlerunning{MoViDNN}

\author{Ekrem \c{C}etinkaya\orcidID{0000-0002-6084-6249} \and
Minh Nguyen\orcidID{0000-0002-9691-1719} \and
Christian Timmerer\orcidID{0000-0002-0031-5243}}
\authorrunning{E. \c{C}etinkaya, M. Nguyen, et al.}

\institute{{Christian Doppler Laboratory ATHENA},
Institute of Information Technology, Alpen-Adria-Universität Klagenfurt, Austria \\
\email{\{firstname.lastname\}@aau.at}}
\maketitle 
\begin{abstract}
Deep neural network (DNN) based approaches have been intensively studied to improve video quality thanks to their fast advancement in recent years. These approaches are designed mainly for desktop devices due to their high computational cost. However, with the increasing performance of mobile devices in recent years, it became possible to execute DNN based approaches in mobile devices. Despite having the required computational power, utilizing DNNs to improve the video quality for mobile devices is still an active research area. In this paper, we propose an open-source mobile platform, namely \textbf{MoViDNN}, to evaluate DNN based video quality enhancement methods, such as super-resolution, denoising, and deblocking. Our proposed platform can be used to evaluate the DNN based approaches both objectively and subjectively. For objective evaluation, we report common metrics such as execution time, PSNR, and SSIM. For subjective evaluation, Mean Score Opinion (MOS) is reported. The proposed platform is available publicly at \url{https://github.com/cd-athena/MoViDNN}.

\end{abstract}
\section{Introduction}
\label{sec:introduction}
The computational power of mobile devices has increased significantly in recent years. With the increasing RAM capacity, CPU power, and, more importantly, the introduction of powerful GPUs, mobile devices have become powerful enough to execute complex tasks, which can only be done with stationary computers until recently. This steep improvement in mobile devices has also increased the number of studies that focus on utilizing deep neural networks in mobile devices~\cite{CVPRWorkshop,AIBenchmark,EVSRNet,liu2021splitsr,PatDNN}.

Video content has become predominant in mobile data traffic. It is estimated to occupy 77\% by 2026~\cite{ericsson_forcast_2026}. Moreover, mobile devices are being mainly used for watching online video content. For example, more than 70\% of watch time on YouTube are from mobile devices~\cite{youtube_mobile}. However, video streaming through mobile broadband suffered unstable and low quality due to severe throughput fluctuations~\cite{trace3G}.

With the increasing number of studies for improving video quality in mobile devices using DNNs, tools to evaluate proposed methods have become an important issue. In this paper, we propose \textbf{MoViDNN}, A \textbf{Mo}bile Platform for Evaluating \textbf{Vi}deo Quality Enhancement with \textbf{D}eep \textbf{N}eural \textbf{N}etworks. The contribution of this paper is two-fold:

\begin{enumerate}[(i)]
    \item \textbf{DNN Based Video Quality Enhancement Evaluation Platform for Mobile Devices:} We provide a platform to apply different machine learning-based approaches to improve video quality and examine their performance in mobile devices with some key objective metrics, including Peak Signal-to-Noise Ratio (PSNR), Structural Similarity Index Measure (SSIM), and execution time. To achieve general results, we apply those approaches for videos belonging to various categories such as animation, sport, movie, \etc. 
    \item \textbf{Subjective Test Platform:} We design a subjective test environment to subjectively evaluate how much improvement in the video quality is perceived by the viewer. The DNN-applied and original videos in the \textbf{DNN test} are shown one-by-one randomly on the screen. The viewer is supposed to assess the quality of the videos on a scale of 1 (bad) to 5 (excellent).
\end{enumerate}

The remainder of this paper is organized as follows. Section~\ref{sec:relatedWork} provides an overview of related work. Section~\ref{sec:app} describes the overall structure and capabilities of our proposed app, followed by implementation details in Section~\ref{sec:implementation}. Finally, conclusions and future work are given in Section~\ref{sec:conclusions}.

\section{Related Work}
\label{sec:relatedWork}
Numerous DNN based enhancement techniques are proposed for images and videos for super-resolution, denoising, and deblocking tasks. 

Shi~\etal~\cite{ESPCN} proposes ESPCN, which utilizes a sub-pixel convolution layer to provide a real-time image and video super-resolution method. In ESPCN, all intermediate feature layers are extracted in the low-resolution (LR) space. In the last layer, final LR features are upscaled with the sub-pixel convolution layer in which an array of upscaling filters are learned.

Liu~\etal~\cite{EVSRNet} proposed EVSRNet for the Mobile AI 2021 challenge~\cite{CVPRWorkshop}. The proposed network consists of residual blocks~\cite{Residual} and a sub-pixel convolution layer which is used at the end to upscale the input. Since the execution time of the network was essential for the challenge, they used neural architecture search (NAS) to determine the optimal hyperparameters for the network. 

Zhang~\etal~\cite{DNCNN} propose DnCNN, which utilizes residual connections and a CNN for image denoising. DnCNN was the first CNN-based denoising approach that can outperform traditional methods. A single model can be used for different applications such as JPEG deblocking and with varying degradation factors. 

Evaluation of DNN performance for mobile devices has gotten attention in recent years thanks to the steep improvement in mobile device capabilities~\cite{AIBenchmark}. 

AI Benchmark Mobile~\cite{AIBenchmark}\footnote{\url{https://ai-benchmark.com/download.html}} is a platform that is proposed to evaluate the performance of mobile devices for executing DNNs. It provides various computer vision tests, including object recognition, image deblurring, and image super-resolution. AI benchmark focuses on the performance of mobile devices for the execution of DNNs instead of evaluating the performance of proposed DNN methods. Therefore, it is not a suitable platform for assessing DNN based video enhancement methods for mobile applications. Moreover, subjective testing is not possible to do with this platform.

Liu~\etal~\cite{liu2021splitsr} investigated an mobile application called ZoomSR for evaluating deep learning-based, on-device SR networks. ZoomSR comprises of two tasks for the participants: 
\begin{inparaenum}[\itshape (i)\upshape]
    \item image task, and
    \item reading task.
\end{inparaenum}
In the former task, the tested images are examined in terms of their quality. In the latter task, the tested images are assessed how hard their text is to be read. Both tasks use a 7-point Likert scale from 1 to 7. Different from this work, our app focuses on evaluating the machine learning-based video enhancement networks in improving video quality and the assessment scale if from 1 to 5, which is commonly used in video quality assessment~\cite{tran2020open,itut,song2010user}.

\section{MoViDNN: A Mobile Platform for Evaluating Video Quality Enhancement with Deep Neural Networks}
\label{sec:app}
The proposed platform is implemented as an Android application and is described in this section.

\subsection{Application Structure}

\begin{figure}[t]
    \centering
    \includegraphics[width=\linewidth]{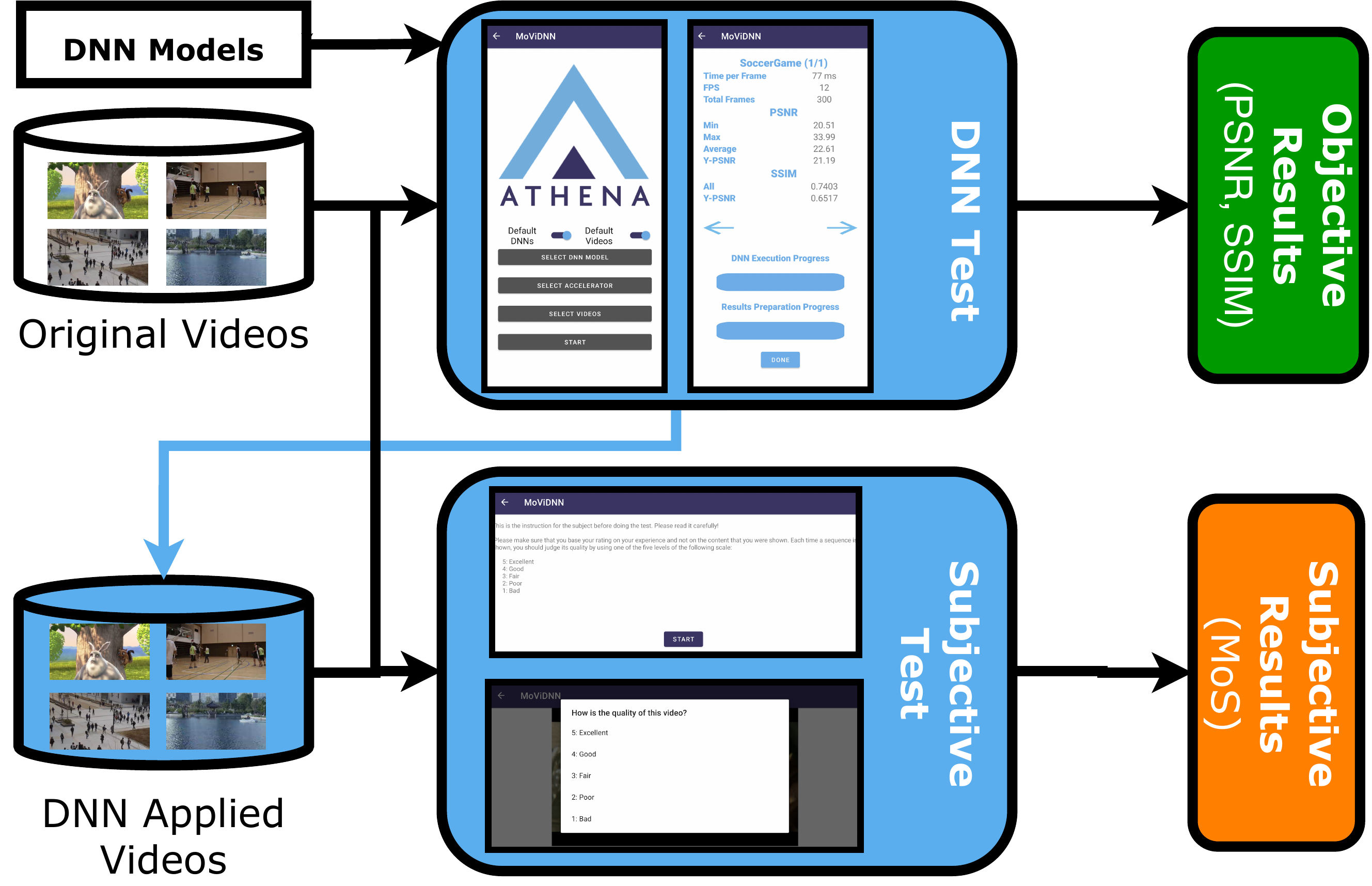}
    \caption{MoViDNN architecture.}
    \label{fig:architecture}
\end{figure}

Fig.~\ref{fig:architecture} shows the MoViDNN architecture comprising two main components:
\begin{inparaenum}[\upshape (a) \itshape]
    \item \textit{DNN Test}, and
    \item \textit{Subjective Test}.
\end{inparaenum}
The \textit{DNN Test} component is responsible for applying DNNs for the input videos from \textit{Original Videos} and save the resulting videos into \textit{DNN-applied Videos}. Moreover, a CSV (comma-separated values) file of corresponding objective metrics (\ie execution time, PSNR, and SSIM) is saved.
The \textit{Subjective Test} component provides an environment to evaluate the quality of videos subjectively by asking the viewers to rate their experience of watching those videos to get the Mean Opinion Score (MOS) and saving MOS in \textit{Results}.

\subsection{DNN Test}
The DNN test of the proposed platform is used to evaluate DNN based video enhancement approaches (\ie super-resolution, denoising, and deblocking) for mobile devices.

Due to the limited computing power of mobile devices, it is essential to apply quantization for the DNN models before using them, as the DNNs are computationally expensive approaches. By applying quantization, it is possible to reduce both the DNN model size and the running time, thus, making them suitable for mobile platforms.

Since 10 seconds of video is suitable for subjective testing~\cite{song2010user}, we limit the length of videos to 10 seconds to reduce the processing on the mobile device. Frames from the input video are extracted in the display order and stored in the local storage temporarily. Afterward, each frame is passed to the DNN in the same order, and the output frame is saved temporarily again. Finally, the resulting frames are concatenated together and converted to a video for the subjective test.

In the final screen of the DNN test, the objective metrics for the given video are displayed. We report PSNR (\textit{minimum, maximum, average} and \textit{y-PSNR)}, SSIM \textit{(all} and \textit{y-SSIM)}, and DNN execution time \textit{(time per frame in milliseconds, frames per second (FPS),} and \textit{total frame count)}.

\subsection{Subjective Test}
The proposed application provides a platform for a subjective test to examine deep learning-based SR and deblocking networks. The subject watches multiple videos, including original and DNN-applied ones generated from the DNN test session. The videos are played in a $1920\times1080$ pixel (Full HD) area. Our application randomly selects either DNN-applied videos or original ones. When each video is played completely, the subject is asked to rate the experience of watching that video using one of the five levels of the following scale:
\begin{inparaenum}[\upshape (1)]
    \item Bad,
    \item Poor,
    \item Fair,
    \item Good,
    \item Excellent.
\end{inparaenum}
The scores of the tested videos in every single subjective test are stored in a CSV file.

\section{Implementation Details}
\label{sec:implementation}
The proposed platform is implemented as a standalone Android application. Tensorflow-lite~\cite{Tensorflow} is used as the DNN framework in the application.

We provide a Github repository \footnote{\url{https://github.com/cd-athena/MoViDNN/tree/main/TFLite_Quantization}} that can be used to convert existing DNN models to mobile compatible versions using Tensorflow~\cite{Tensorflow} as the backend. Once a network is quantized and converted to a \textit{tensorflow-lite} version, it can be evaluated in the platform.

The proposed application comes with three DNN models, namely, ESPCN~\cite{ESPCN}, EVSRNet~\cite{EVSRNet}, and DnCNN~\cite{DNCNN}. Multiple videos in~\cite{li2019avc} with different content genres (\eg movie, sports, natural scene, architecture) are included as well. MoViDNN will automatically update the list of available networks and videos once additions are made.

\section{Conclusions and Future Work}
\label{sec:conclusions}
In this paper, we introduced \textbf{MoViDNN} to evaluate DNN-based video quality enhancement methods in mobile devices. \textbf{MoViDNN} can be used for both objective and subjective evaluation. For objective evaluation, PSNR, SSIM, and execution time are calculated and reported. For subjective evaluation, MOS is calculated for each test video. We include several state-of-the-art DNNs in the context of super-resolution and deblocking in the demonstration. Moreover, we provide a Github repository that can be used to convert and evaluate additional DNNs with ~\textbf{MoViDNN}. Finally, a real-world demonstration of the ~\textbf{MoViDNN} can be seen in the provided video~\footnote{\url{https://www.youtube.com/watch?v=MzeEsNRlVv0}}.

As future work, we plan to improve the subjective test part of ~\textbf{MoViDNN} by including a crowdsourcing option. Moreover, extending the platform to support additional DNN based video quality enhancement methods such as video frame interpolation can also be done.

\section*{Acknowledgment}
\small{The financial support of the Austrian Federal Ministry for Digital and Economic Affairs, the National Foundation for Research, Technology and Development, and the Christian Doppler Research Association, is gratefully acknowledged. Christian Doppler Laboratory ATHENA: \url{https://athena.itec.aau.at/}.}

\bibliographystyle{splncs04}
\scriptsize{\bibliography{ref}}

\normalsize
\appendix
\pagebreak
\section{Running the Application}
After installing the application, make sure you give the permission of allowing management of all files.
\subsection{DNN Test}
Steps for evaluating a new DNN structure are as following:
\begin{enumerate}
    \item Use the Github repository to quantize and convert the DNN model into a \textit{tensorflow-lite} model.
    \item Place the quantized network into the allocated folder in the mobile device storage.
    \item Place the desired input video to the allocated folder in the mobile device storage. This step is optional as we provide several test sequences in the application by default.
    \item Click on \textbf{DNN TEST} button in the home screen (see Fig.~\ref{fig:homescreen}).
    \item Pick the DNN model, the accelerator (CPU, GPU, NNAPI), and the videos in the DNN configuration page of the application and start the evaluation process as illustrated in Fig.~\ref{fig:dnnconfig}.
    \item Wait for the process to be completed, which will take some time (Fig.~\ref{fig:dnnprocessing}).
    \item Once the DNN test is completed as in Fig.~\ref{fig:dnncomplete}, a subjective test can be run for the new DNN structure.
\end{enumerate}

\subsection{Subjective test}
\begin{figure}
    \begin{subfigure}{0.22\linewidth}
        \centering
        \includegraphics[width=\linewidth]{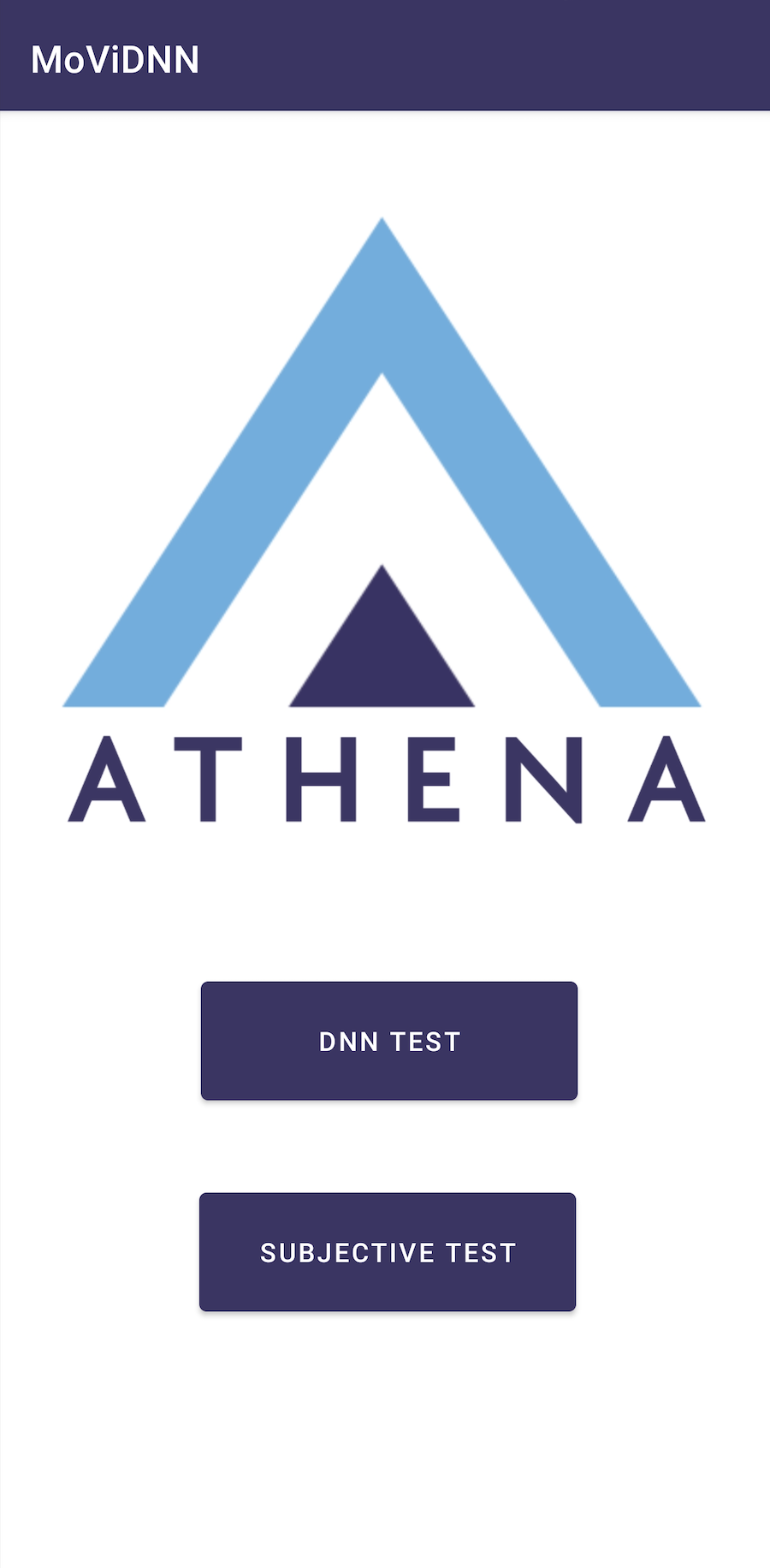}
        \caption{}
        \label{fig:homescreen}
    \end{subfigure}
    \begin{subfigure}{0.22\linewidth}
        \centering
        \includegraphics[width=\linewidth]{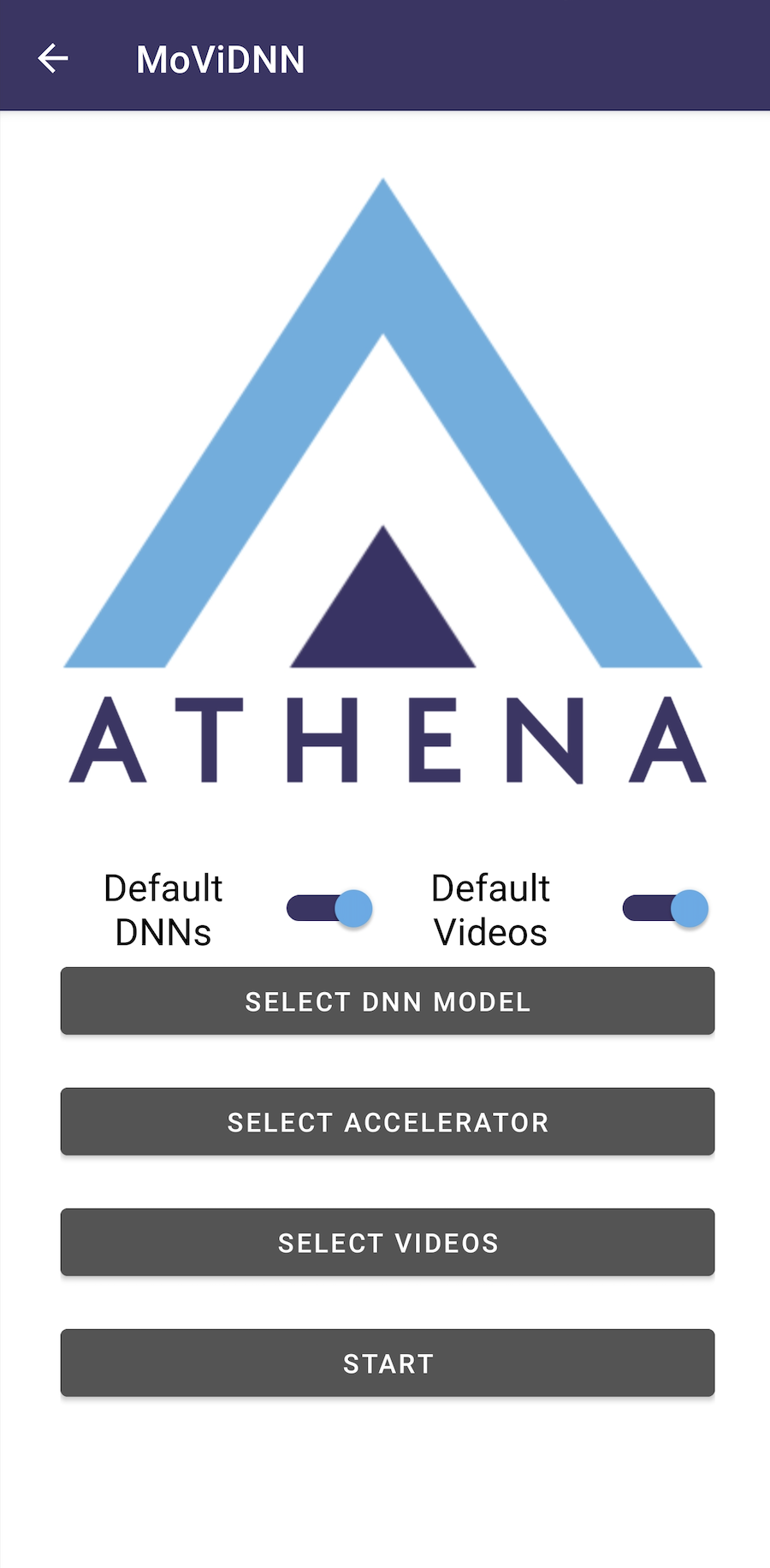}
        \caption{}
        \label{fig:dnnconfig}
    \end{subfigure}
    \begin{subfigure}{0.22\linewidth}
        \centering
        \includegraphics[width=\linewidth]{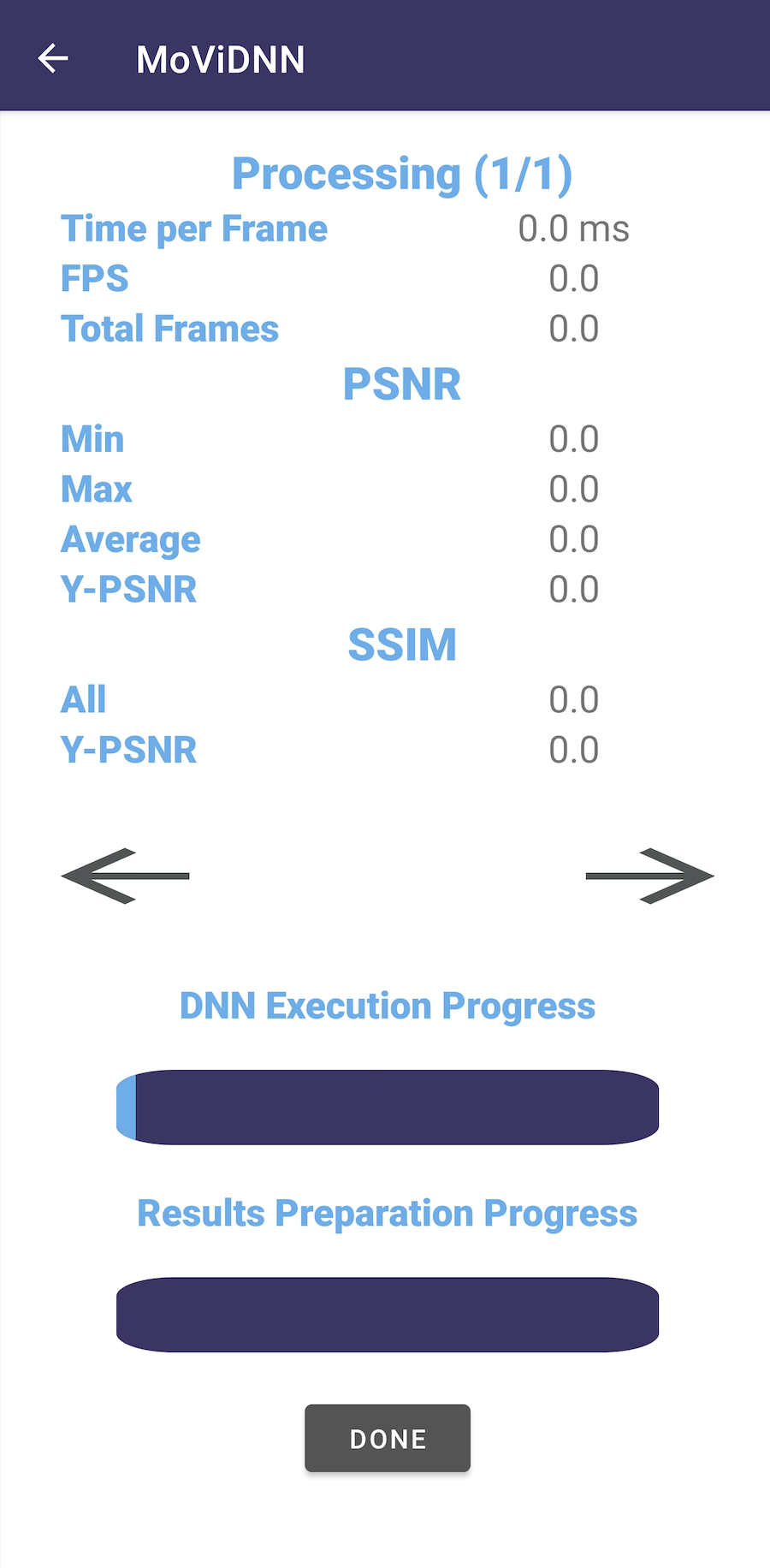}
        \caption{}
        \label{fig:dnnprocessing}
    \end{subfigure}
    \begin{subfigure}{0.22\linewidth}
        \centering
        \includegraphics[width=\linewidth]{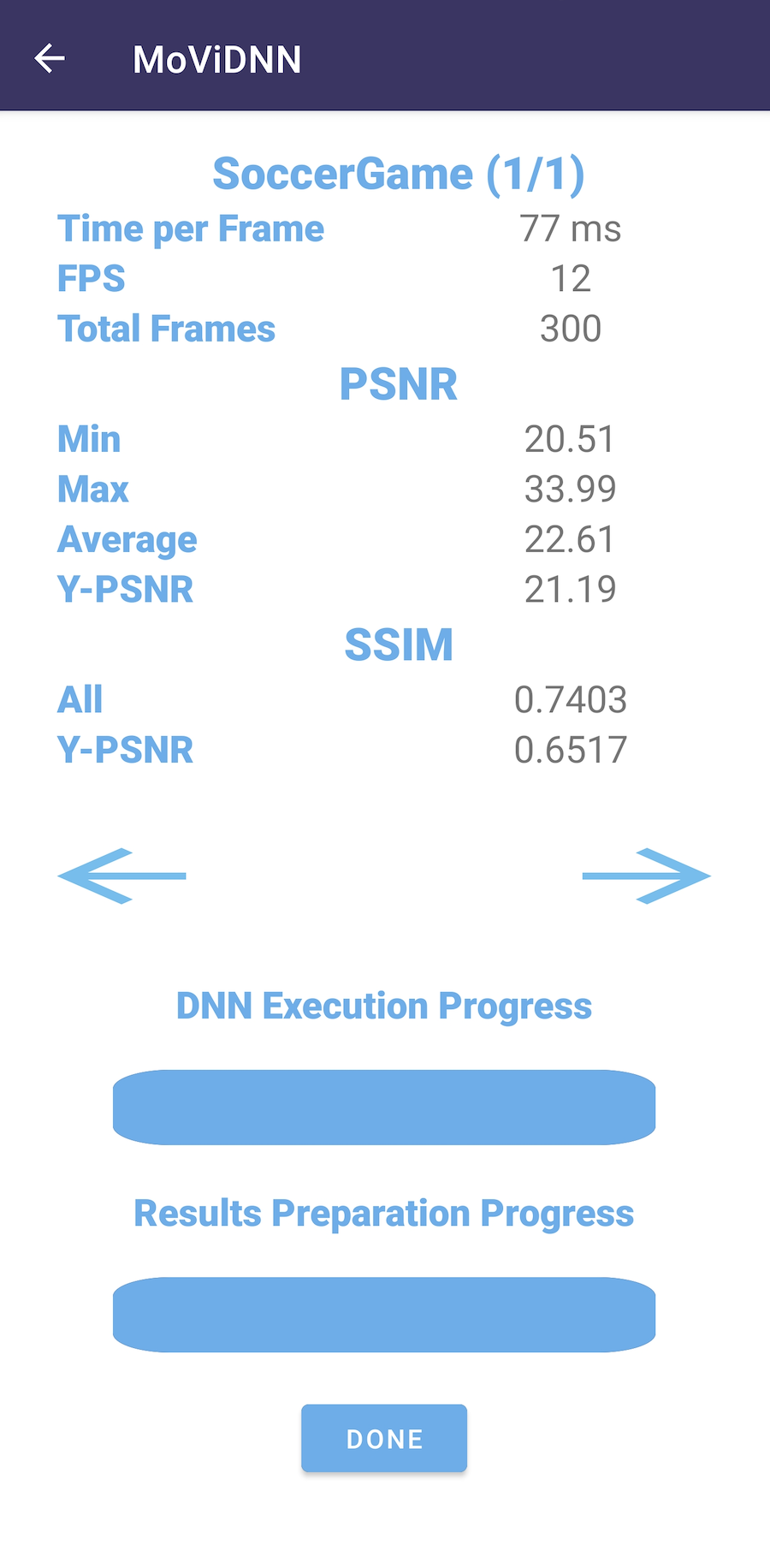}
        \caption{}
        \label{fig:dnncomplete}
    \end{subfigure}
    \caption{DNN test UI of MOVIDNN. \textit{(a)} Home screen, \textit{(b)} Network, accelerator and video selection, \textit{(c)} Processing, \textit{(d)} Complete process.}
\end{figure}

\begin{figure}
    \begin{subfigure}{0.22\linewidth}
        \centering
        \includegraphics[width=\linewidth]{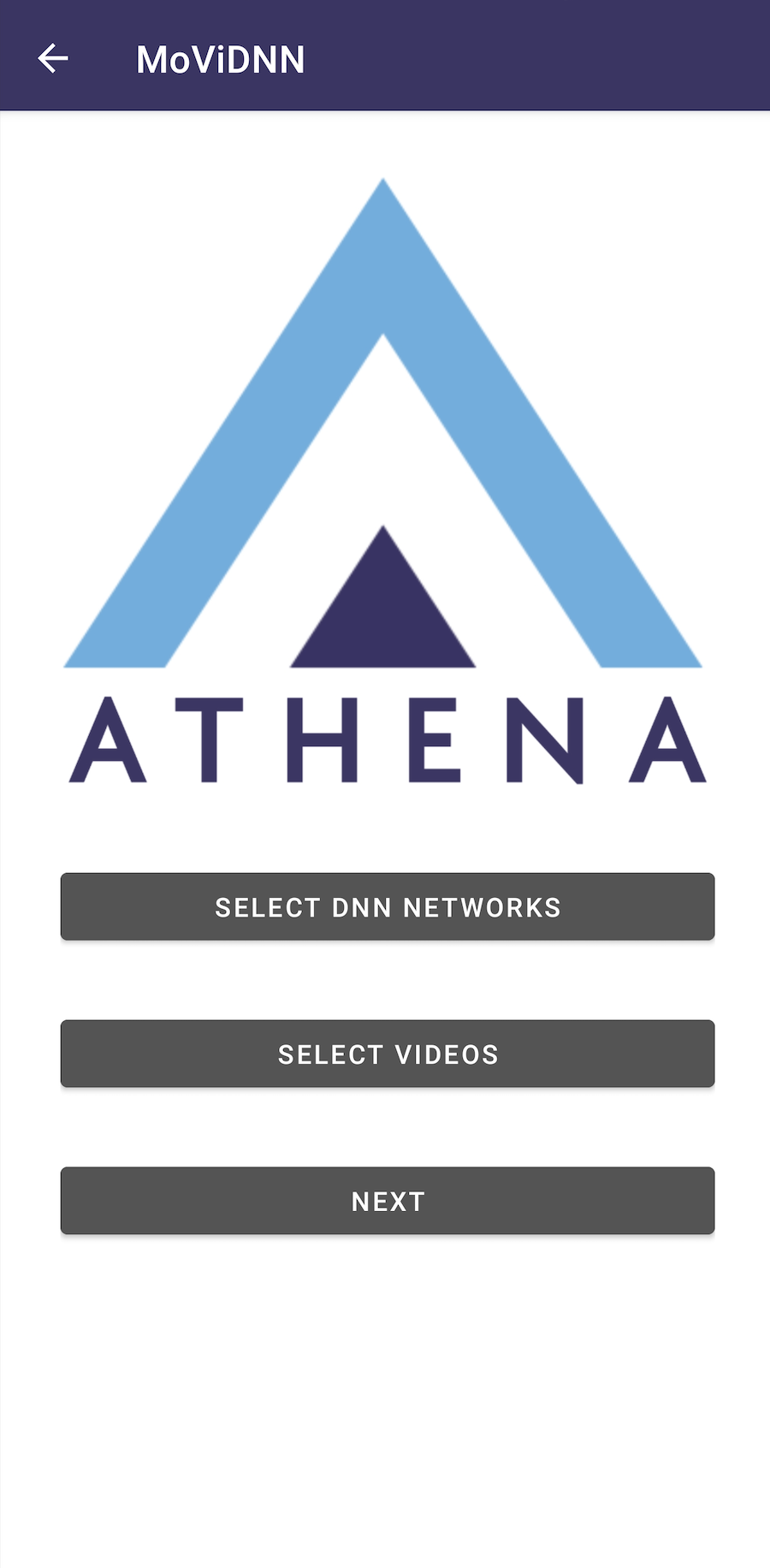}
        \caption{}
        \label{fig:subjectiveNetwork}
    \end{subfigure}
    \begin{subfigure}{0.22\linewidth}
        \centering
        \includegraphics[width=\linewidth]{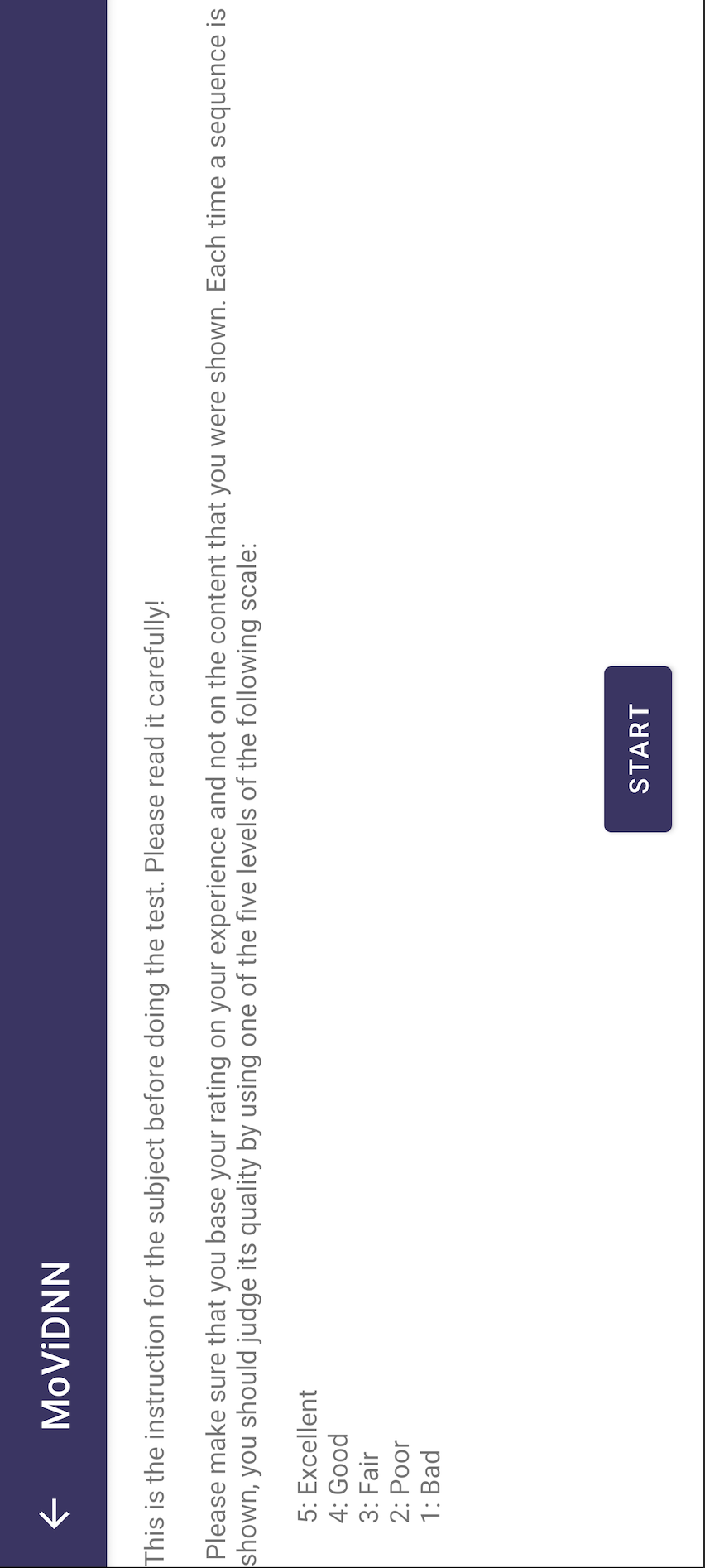}
        \caption{}
        \label{fig:subjectiveInstruction}
    \end{subfigure}
    \begin{subfigure}{0.22\linewidth}
        \centering
        \includegraphics[width=\linewidth]{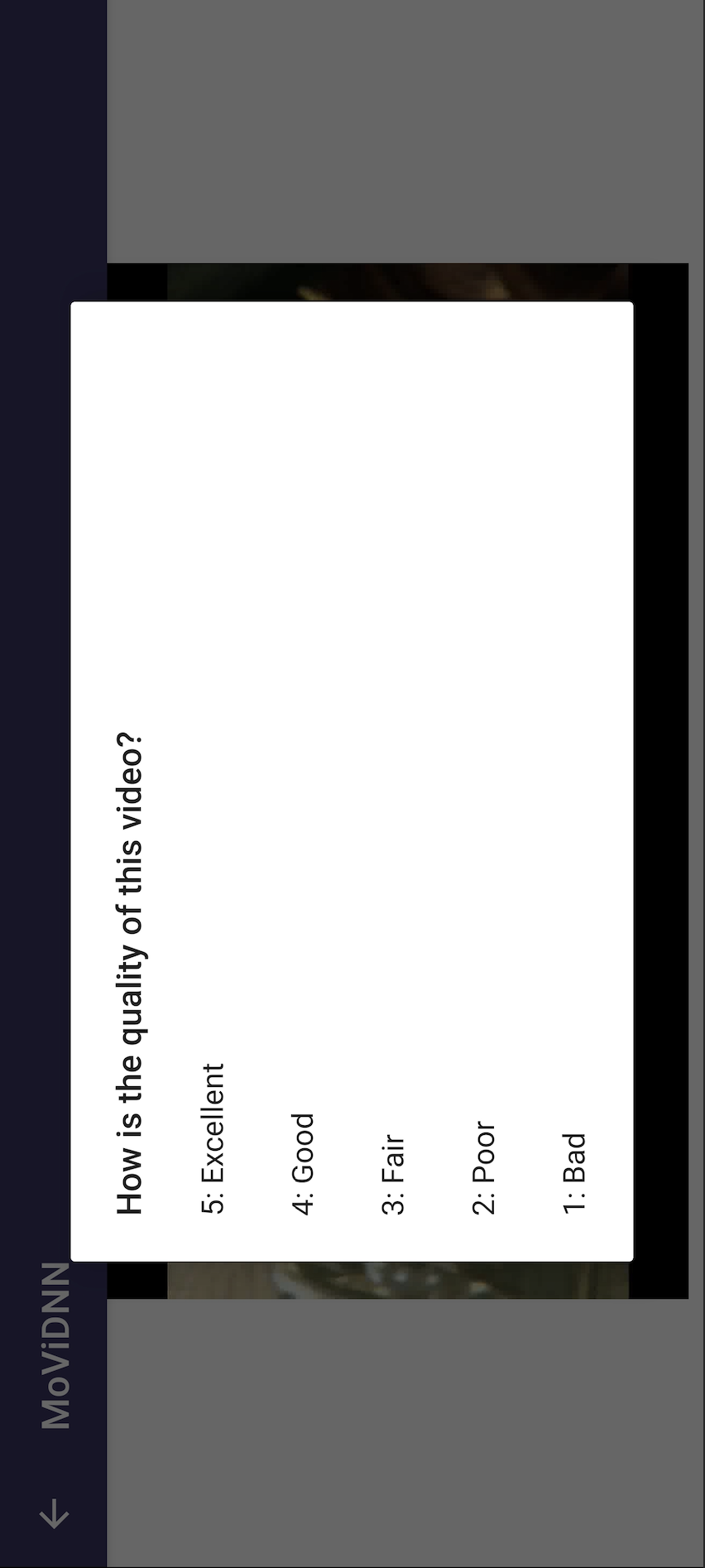}
        \caption{}
        \label{fig:subjectiveRate}
    \end{subfigure}
    \begin{subfigure}{0.22\linewidth}
        \centering
        \includegraphics[width=\linewidth]{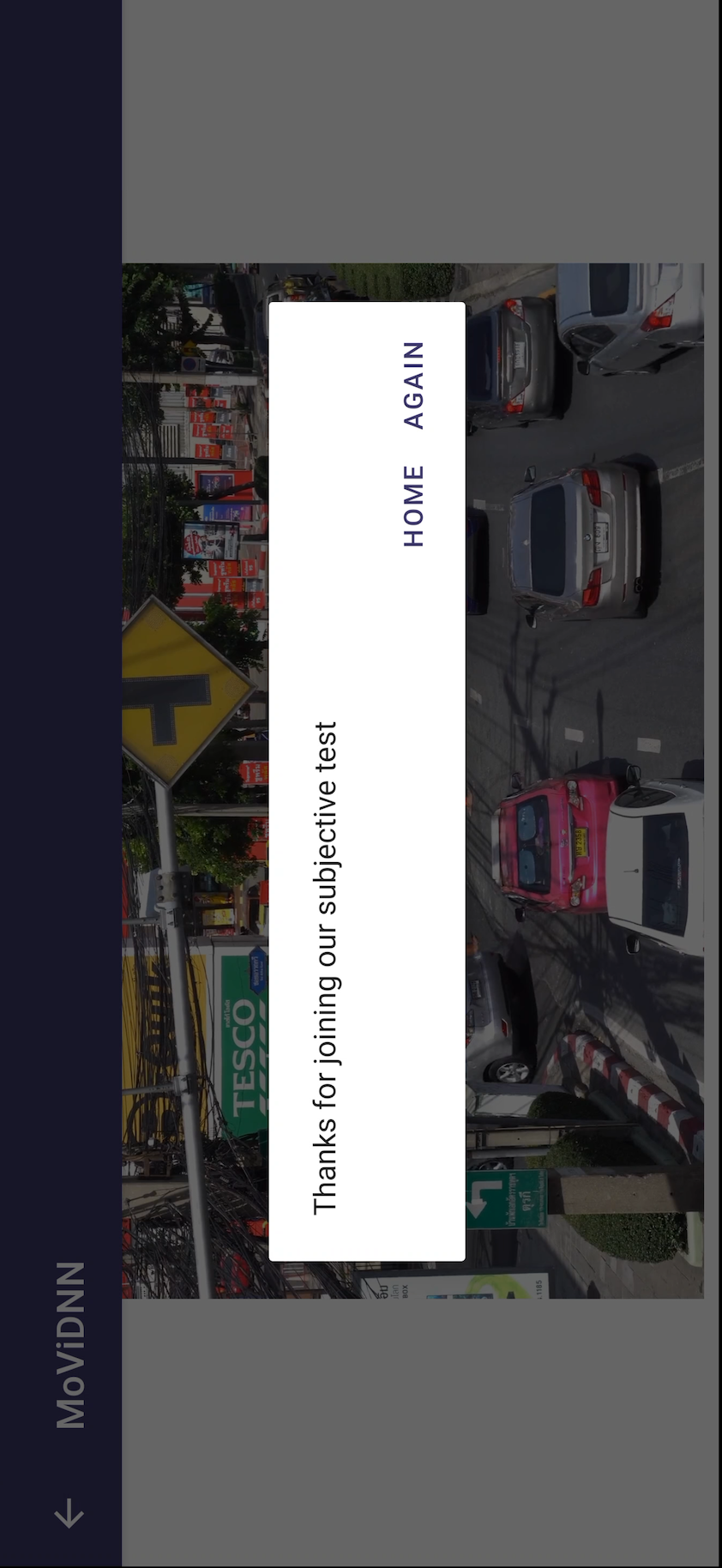}
        \caption{}
        \label{fig:subjectiveEnd}
    \end{subfigure}
    \caption{Subjective test UI of MOVIDNN. \textit{(a)} Network and video selection, \textit{(b)} Subjective Instruction, \textit{(c)} Video assessment, \textit{(d)} End of a test.}
\end{figure}

Steps for running a subjective test are as following:
\begin{enumerate}
    \item Click on \textbf{SUBJECTIVE TEST} button in the home screen (see Fig.~\ref{fig:homescreen}).
    \item Pick the DNN models and the videos in the Subjective configuration page of the application as illustrated in Fig.~\ref{fig:subjectiveNetwork}. Click \textbf{NEXT} button.
    \item Read carefully the instruction of subjective test about how to rate the experience of watching the video. Click \textbf{START} button (see Fig.~\ref{fig:subjectiveInstruction}).
    \item To play a new video, click \textbf{CONTINUE} button and watch the whole video. Rate the quality of the video when a pop-up is shown. Here, 5 means excellent, and 1 means bad. The pop-up screen is shown in Fig.~\ref{fig:subjectiveRate}.
    \item Repeat the previous step for other videos until the Subjective test is finished.
    \item Click \textbf{HOME} button to return to the home screen of the app or \textbf{AGAIN} button for another participant doing the subjective test (see Fig.~\ref{fig:subjectiveEnd}).
\end{enumerate}

\end{document}